\documentclass[10pt,twocolumn,letterpaper]{article}

\usepackage{cvpr}              %

\usepackage{graphicx}
\usepackage{amsmath}
\usepackage{amssymb}
\usepackage{booktabs}
\usepackage{multicol}
\usepackage{multirow}
\usepackage{pifont}
\usepackage{adjustbox}
\usepackage{arydshln}
\usepackage{colortbl}
\definecolor{cssgreen}{rgb}{0.0, 0.5, 0.0}
\definecolor{cssred}{rgb}{1, 0, 0.0}
\definecolor{DarkGreen}{rgb}{0.43, 0.68, 0.28}

\usepackage{colortbl}
\usepackage{booktabs}
\usepackage{multirow}
\definecolor{greenx}{HTML}{229954}
\definecolor{aliceblue}{rgb}{0.94, 0.97, 1.0}
\definecolor{f2ecde}{HTML}{f2ecde}
\definecolor{alizarin}{rgb}{0.82, 0.1, 0.26}
\definecolor{royalblue}{RGB}{65,105,225}

\usepackage{adjustbox}
\usepackage{marvosym}
\usepackage{tikz}

\newcommand{\ourmethod}{DEIM}

\definecolor{cvprblue}{rgb}{0.21,0.49,0.74}
\usepackage[pagebackref,breaklinks,colorlinks,allcolors=cvprblue]{hyperref}

\title{Real-Time Object Detection Meets DINOv3}

\author{
Shihua Huang\textsuperscript{1\(\star\)}, 
Yongjie Hou\textsuperscript{1, 2\(\star\)}, 
Longfei Liu\textsuperscript{1\(\star\)}, 
Xuanlong Yu\textsuperscript{1}, 
Xi Shen\textsuperscript{1\(\dagger\)} \\[0.5em]
\textsuperscript{1} Intellindust AI Lab; 
\textsuperscript{2}Xiamen University
\\
\textsuperscript{\(\star\)} Equal Contribution; \textsuperscript{\(\dagger\)} Corresponding author.
\\[0.5em]
Project Page: \href{https://intellindust-ai-lab.github.io/projects/DEIMv2/}{ https://intellindust-ai-lab.github.io/projects/DEIMv2}
\\
Code \& Weights: \href{https://github.com/Intellindust-AI-Lab/DEIMv2}{ https://github.com/Intellindust-AI-Lab/DEIMv2}
}

\begin{document}
\setcounter{figure}{-1}

\twocolumn[{
\maketitle
\begin{center}
\vspace{-0.2cm}
    \captionsetup{type=figure}
    \hfill
    \begin{subfigure}{0.494\textwidth}
\includegraphics[width=0.97\textwidth]{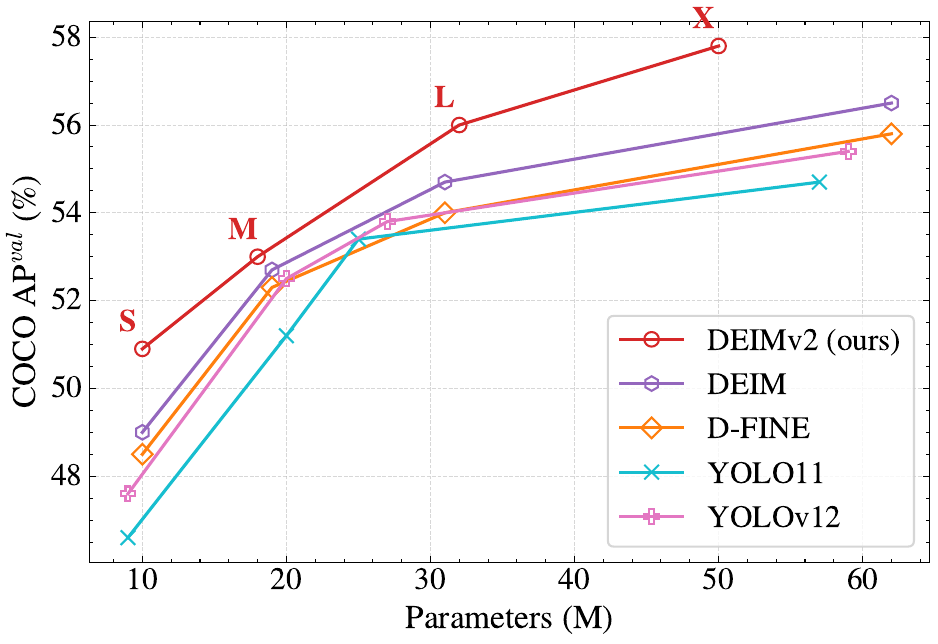}
        \caption{\textbf{COCO performance v.s. Number of Parameters.}}
        
    \end{subfigure}
    \hfill
    \begin{subfigure}{0.494\textwidth}
        \includegraphics[width=0.97\textwidth]{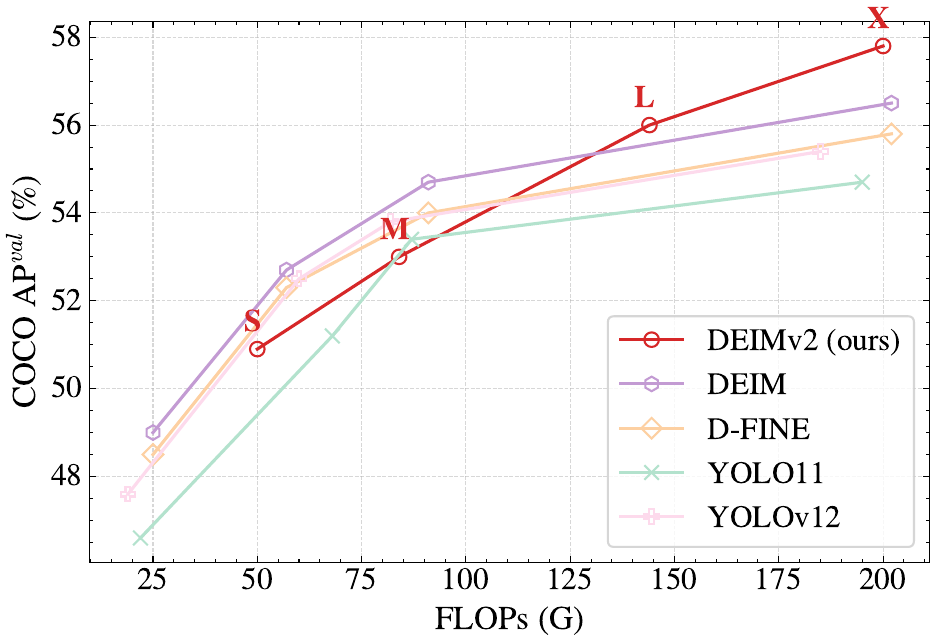}
        \caption{\textbf{COCO performance v.s. FLOPs.}}
        
    \end{subfigure}
    \hfill 
    \vspace{-0.2cm}
    \captionof{figure}{Compared with state-of-the-art real-time object detectors on COCO \cite{lin2014microsoft}, all variants of our proposed DEIMv2 (S, M, L, X) achieve superior performance in terms of average precision (AP) while maintaining similar parameters.}
    \label{fig:teaser}
\end{center}
}]

\begin{abstract}
   Driven by the simple and effective Dense O2O, DEIM demonstrates faster convergence and enhanced performance. In this work, we extend it with DINOv3 features, resulting in DEIMv2. DEIMv2 spans eight model sizes from X to Atto, covering GPU, edge, and mobile deployment. For the X, L, M, and S variants, we adopt DINOv3-pretrained / distilled backbones and introduce a Spatial Tuning Adapter (STA), which efficiently converts DINOv3’s single-scale output into multi-scale features and complements strong semantics with fine-grained details to enhance detection. For ultra-lightweight models (Nano, Pico, Femto, and Atto), we employ HGNetv2 with depth and width pruning to meet strict resource budgets. Together with a simplified decoder and an upgraded Dense O2O, this unified design enables DEIMv2 to achieve a superior performance–cost trade-off across diverse scenarios, establishing new state-of-the-art results.
   Notably, our largest model, DEIMv2-X, achieves 57.8 AP with only 50.3M parameters, surpassing prior X-scale models that require over 60M parameters for just 56.5 AP. On the compact side, DEIMv2-S is the first sub-10M model (9.71M) to exceed the 50 AP milestone on COCO, reaching 50.9 AP. Even the ultra-lightweight DEIMv2-Pico, with just 1.5M parameters, delivers 38.5 AP—matching YOLOv10-Nano (2.3M) with $\sim$50\% fewer parameters. %
\end{abstract}
    
\section{Introduction}

\begin{table*}[t]
    \footnotesize
    \centering
    \renewcommand{\arraystretch}{1.05}
    \setlength{\tabcolsep}{5.5pt}
    \caption{
        \textbf{Architectural details and main results of DEIMv2 variants.} We report the configurations for different variants and the final Average Precision (AP) on the COCO benchmark.
    }
    \vspace*{-2mm}
    \begin{tabular*}{\textwidth}{l | lcccccccc | ccc | c}
        \toprule
        \hline
        \multirow{2}{*}{\textbf{Variant}} & \multicolumn{2}{c}{\textbf{Backbone}} & \multicolumn{3}{c}{\textbf{Hidden Dimension}} & \multicolumn{2}{c}{\textbf{Layers}}  & \multirow{2}{*}{\textbf{\#Scales}} & \multirow{2}{*}{\textbf{\#Query}}  & \multirow{2}{*}{\textbf{\#Params}} & \multirow{2}{*}{\textbf{GFLOPs}} & \multirow{2}{*}{\textbf{Latency}} & \multirow{2}{*}{\textbf{AP}} \\
        & \textbf{Model} & \textbf{Adapter} & $\mathbf{d_{Back.}}$ & $\mathbf{d_{Enc.}}$ & $\mathbf{d_{Dec.}}$ & \textbf{\#$_{\text{Back.}}$} & \textbf{\#$_{\text{Dec.}}$} & & & & & & \\
        \midrule
        \hline
        X & ViT-S+ & STA & 384 & 256 & 256 & 12 & 6 & 3 & 300 & 50.26 & 200.0 & 13.75 & 57.8 \\
        L & ViT-S & STA & 384 & 256 & 256 & 12 & 4 & 3 & 300 & 32.18 & 144.5 & 10.47 & 56.0 \\
        M & ViT-T+ & STA & 256 & 256 & 256 & 12 & 4 & 3 & 300 & 18.11 & 84.1 & 8.80 & 53.0 \\
        S & ViT-T  & STA & 192 & 192 & 192 & 12 & 4 & 3 & 300 & 9.71 & 49.5 & 5.78 & 50.9 \\
        \midrule
        Nano & HGv2-B0 & - & 1024 & 128 & 128 & 5 & 3 & 2 & 300 & 3.57 & 6.86 & 2.32 & 43.0 \\
        Pico & HGv2-P & - & 512 & 112 & 112 & 4 & 3 & 2 & 200 & 1.51 & 5.15 & 2.14 & 38.5 \\
        Femto & HGv2-F & - & 256 & 96 & 96 & 3 & 3 & 2 & 150 & 0.96 & 1.67 & 1.91 & 31.0 \\
        Atto & HGv2-A & - & 256 & 64 & 64 & 3 & 3 & 2 & 100 & 0.49 & 0.76 & 1.61 & 23.8 \\
        \bottomrule
    \end{tabular*}
    \label{tab:main}
\end{table*}
Real-time object detection~\cite{redmon2016you,yolo11,tian2025yolov12,zhao2024detrs} is a critical component in many practical applications, including autonomous driving~\cite{liang2022edge}, robotics~\cite{maji2024yolo}, and industrial defect detection. Achieving a good balance between detection performance and computational efficiency remains a key challenge, especially for lightweight models suitable for edge and mobile devices. 

Among prevailing real-time detectors, DETR-based methods are increasingly favored for their end-to-end nature. Despite the robust feature representations offered by DINOv3~\cite{simeoni2025dinov3} in diverse scenarios, the feasibility and effectiveness of integrating it into DETR-based models have yet to be fully investigated.

In this work, we introduce DEIMv2, a real-time object detector built upon our previous DEIM~\cite{huang2025deim} pipeline and enhanced with DINOv3~\cite{simeoni2025dinov3} features. DEIMv2 employs official DINOv3-pretrained backbones (ViT-Small and ViT-Small+) for its largest variants (L and X sizes) to maximize feature richness, while its S and M variants leverage ViT-Tiny and ViT-Tiny+ backbones distilled from DINOv3, carefully balancing performance and efficiency. To address ultra-lightweight scenarios, we further introduce four specialized variants: Nano, Pico, Femto, and Atto, extending DEIMv2’s scalability across a wide spectrum of computational budgets.

To better leverage the strong feature representations of DINOv3, pretrained on large-scale data, under real-time constraints, we design the Spatial Tuning Adapter (STA). Operating in parallel with DINOv3, STA efficiently converts its single-scale outputs into the multi-scale features required for object detection in a parameter-free manner. Simultaneously, it performs fast downsampling of the input image to provide fine-grained, multi-scale detail features with very small receptive fields, complementing DINOv3’s strong semantics.

We further simplify the decoder by drawing on advances from the Transformer community. Specifically, we replace the conventional FFN and LayerNorm with SwishFFN~\cite{shazeer2020glu} and RMSNorm~\cite{zhang2019root}, both of which have been shown to be efficient without significantly affecting performance. We additionally note that object query locations change minimally during iterative refinement, motivating sharing query position embeddings across all decoder layers. Beyond this, we enhance Dense O2O by introducing object-level Copy-Blend augmentation, which increases effective supervision and further improves model performance.

\begin{figure*}[t]
    \centering
    \includegraphics[width=\linewidth]{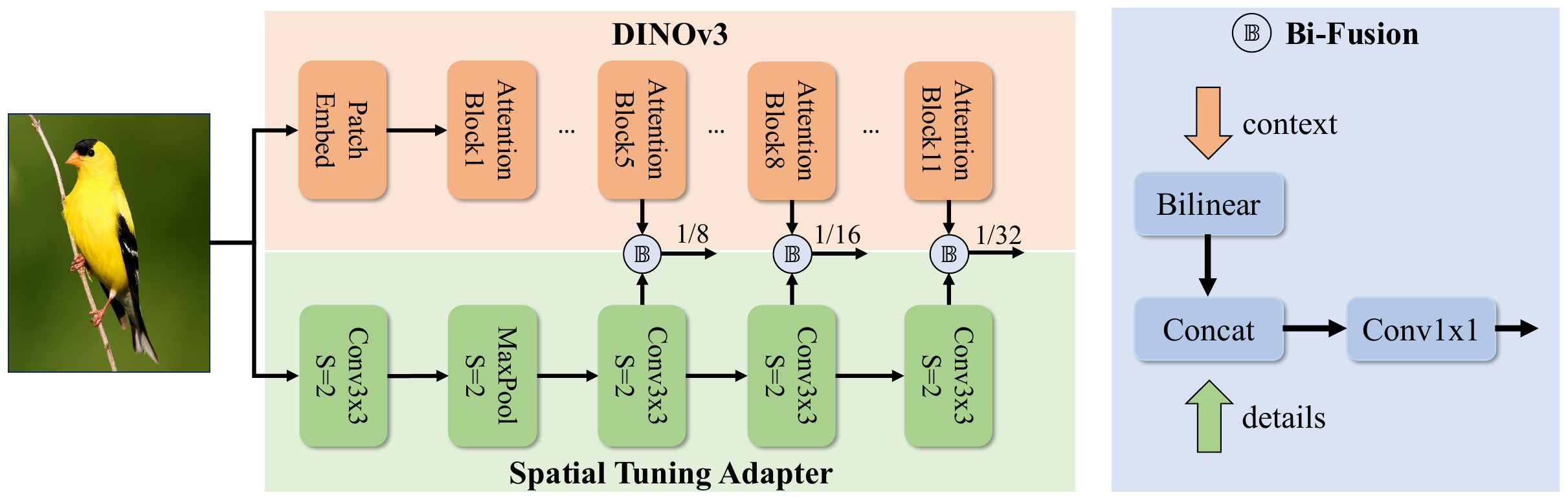}
    \caption{\textbf{Backbone design of our ViT-based variants}. We integrate DINOv3 with the proposed Spatial Tuning Adapter (STA).}
    \label{fig:STA}
\end{figure*}

Extensive experiments on COCO~\cite{lin2014microsoft} demonstrate that DEIMv2 achieves state-of-the-art performance across multiple model scales, which can be seen in Figure~\ref{fig:teaser}. Despite its simplicity, the family of DEIMv2 exhibits strong performance. For instance, our largest variant, DEIMv2-X, achieves 57.6 AP on COCO with only 50.3M parameters, surpassing prior best X-scale detectors DEIM-X that requires over 60M parameters yet attains only 56.5 AP. At the smaller end, DEIMv2-S establishes a notable milestone as the first model with fewer than 10M parameters to exceed 50 AP, highlighting the effectiveness of our design at compact scales. Furthermore, our ultra-lightweight DEIMv2-Pico, with merely 1.5M parameters, attains 38.5 AP, matching the performance of YOLOv10-Nano (2.3M parameters) while reducing parameter count by $\sim$50\%, thereby redefining the efficiency–accuracy frontier at the extreme lightweight regime. 

Our work highlights how DINOv3~\cite{simeoni2025dinov3} features can be effectively adapted for real-time object detection and provides a versatile framework spanning ultra-lightweight to high-performance models. To our knowledge, this is the first work in real-time object detection to simultaneously address such a wide range of deployment scenarios.

The main contributions of this work are summarized as follows:

\begin{itemize}

\item We present DEIMv2, which offers eight model sizes covering GPU, edge, and mobile deployment.

\item For larger models, we leverage DINOv3 for strong semantic features and introduce the STA to efficiently integrate them into real-time object detection.

\item For ultra-lightweight models, we leverage expert knowledge to effectively prune the depth and width of HGNetv2-B0, meeting strict computational constraints. 

\item Beyond backbone, we further simplify the decoder and upgrade Dense O2O, pushing the performance boundaries even further.

\item Finally, we demonstrate on COCO that DEIMv2 outperforms existing state-of-the-art methods across all resource settings, establishing new SOTA results.

\end{itemize}

\section{Method}

\paragraph{Overall architecture.} Our overall architecture follows the design of RT-DETR~\cite{lv2024rt}, comprising a backbone, a hybrid encoder, and a decoder. As shown in Table~\ref{tab:main}, for the mainstream variants X, L, M, and S, the backbone is based on DINOv3 with our proposed Spatial Tuning Adapter (STA), while the remaining variants use HGNetv2~\cite{cui2021beyond}. Multi-scale features from the backbone are first processed by the encoder to produce initial detection results and select the top-$K$ candidate bounding boxes. The decoder iteratively refines these candidates to generate the final predictions.

\paragraph{ViT-based variants.} For the larger DEIMv2 variants (S, M, L, X), we carefully design the backbones around the Vision Transformer~\cite{dosovitskiy2020image} (ViT) family, balancing model capacity with efficiency. For L and X, we leverage two public DINOv3 models~\cite{simeoni2025dinov3}: ViT-Small and ViT-Small+, which provide strong semantic representations with 12 layers and a 384-dimensional hidden size. For the lighter S and M variants, we distill compact backbones, ViT-Tiny and ViT-Tiny+, directly from ViT-Small DINOv3, preserving the 12-layer depth while reducing the hidden dimensions to 192 and 256. This design delivers a smooth scaling path across S $\rightarrow$ M $\rightarrow$ L $\rightarrow$ X, ensuring that each variant maintains competitive accuracy while adapting to different efficiency requirements.

\paragraph{HGNetv2-based variants.} 
HGNetv2~\cite{cui2021beyond}, developed by the Baidu PaddlePaddle team, is widely used in real-time DETR frameworks for its efficiency—for example, D-FINE~\cite{peng2024d} adopts the full HGNetv2 series as its backbone. In our ultra-lightweight DEIMv2 models (Nano, Pico, Femto, and Atto), we also build on HGNetv2-B0, but progressively prune its depth and width to meet different parameter budgets. Specifically, the Pico backbone removes the fourth stage of B0, keeping outputs only up to 1/16 scale. Femto further reduces the number of blocks in Pico’s last stage from two to one. Atto goes a step further by shrinking the channels of that last block from 512 to 256.

\paragraph{Spatial Tuning Adapter.}
To better adapt DINOv3 features for real-time object detection, we propose the Spatial Tuning Adapter (STA), as illustrated in Fig.~\ref{fig:STA}. STA is a fully convolutional network that integrates an ultra-lightweight feedforward network for extracting fine-grained multi-scale details, together with a Bi-Fusion operator that further strengthens feature representations from DINOv3.

\begin{table*}[h]
\centering
\caption{\textbf{Detailed training hyperparameters in DEIMv2.} Back. denotes the Backbone. We use \textit{Local Loss} to denote the Fine-Grained Localization (FGL) Loss and the Decoupled Distillation Focal (DDF) Loss in D-FINE~\cite{peng2024d}.}
\label{tab:hyperparams}
    \begin{tabular}{l|cccccccc}
        \toprule
        \textbf{HyperParams} & \textbf{X} & \textbf{L} & \textbf{M} & \textbf{S} & \textbf{Nano} & \textbf{Pico} & \textbf{Femto} & \textbf{Atto} \\
        \midrule
        Resolution & 640 & 640 & 640 & 640 & 640 & 640 & 416 & 320 \\
        \midrule
        Weigt Decay & 1.25e-4 & 1.25e-4 & 1e-4 & 1e-4 & 1e-4 & 1e-4 & 1e-4 & 1e-4 \\
        Base LR & 5e-4 & 5e-4 & 5e-4 & 5e-4 & 8e-4 & 1.6e-3 & 1.6e-3 & 2e-3 \\
        Min LR & 2.5e-4 & 2.5e-4 & 2.5e-4 & 2.5e-4 & 8e-4 & 8e-4 & 8e-4 & 1e-3 \\
        Back. LR & 1e-6 & 1.25e-5 & 2.5e-5 & 2.5e-5 & 4e-4 & 8e-4 & 8e-4 & 1e-3 \\
        Back. MinLR & 5e-7 & 6.25e-6 & 1.25e-5 & 1.25e-5 & 4e-4 & 4e-4 & 4e-4 & 5e-4 \\
        \midrule
        Local Loss & \checkmark & \checkmark & \checkmark & \checkmark & \checkmark \\
        Epochs & 50 & 60 & 90 & 120  & 148 & 468 & 468 & 468 \\
        \midrule
        Mosaic$_{prob}$ & 0.5 & 0.5 & 0.5 & 0.5 & 0.5 & 0.5 & 0.5 & 0.3 \\
        Mosaic$_{epochs}$ & [4, 29] & [4, 34] & [4, 49] & [4, 64] & [4, 78] & [4, 250] & [4, 250] & [4, 250] \\
        MixUp$_{prob}$ & 0.5 & 0.5 & 0.5 & 0.5 & 0.5 & 0.0 & 0.0 & 0.0 \\
        MixUp$_{epochs}$ & [4, 29] & [4, 34] & [4, 49] & [4, 64] & [4, 78] & [ ] & [ ] & [ ] \\
        CopyB$_{prob}$ & 0.5 & 0.5 & 0.5 & 0.5 & 0.4 & 0.0 & 0.0 & 0.0  \\
        CopyB$_{epoch}$ & [4, 50] & [4, 60] & [4, 90] & [4, 120] & [4, 78] & [ ] & [ ] & [ ]  \\
        \bottomrule
    \end{tabular}
\end{table*}

DINOv3 is based on a ViT backbone, which naturally produces single-scale (1/16) dense features. In object detection, however, objects vary widely in size, and multi-scale features are one of the most effective ways to improve performance. To this end, ViTDet~\cite{li2022exploring} introduced the Feature2Pyramid module, which generates multi-scale features from the final ViT output using deconvolution. In contrast, our STA is even simpler: we directly resize the 1/16-scale features from several ViT blocks (e.g., the 5th, 8th, and 11th) into multiple scales via parameter-free bilinear interpolation. These multi-scale features are further enhanced by the Bi-Fusion operator consisting of $1\times1$ convolution with an ultra-lightweight CNN designed to extract fine-grained details and complement DINOv3’s output features. This design achieves an excellent trade-off between efficiency and accuracy, making it well-suited for real-time detection.

\begin{table*}[t]
    \caption{\textbf{Comparison with real-time object detectors on COCO~\cite{lin2014microsoft} \texttt{val2017}, sorted by parameter size.} }
    \label{tab:XLMS}
    \begin{center}
    \begin{adjustbox}{width=\textwidth}
    \begin{tabular}{l | cccc | ccc ccc}
        \toprule
        \hline
        Model & \#Epochs & \#Params. & GFLOPs & Latency (ms) & AP$^{val}$ & AP$^{val}_{50}$ & AP$^{val}_{75}$ & AP$^{val}_S$ & AP$^{val}_M$ & AP$^{val}_L$ \\
        \midrule
        \hline
        YOLOv10-S~\cite{wang2024yolov10} & 500 & 7 & 22 & 2.52 & 46.3 & 63.0 & 50.4 & 26.8 & 51.0 & 63.8 \\
        YOLOv9-S~\cite{wang2024yolov9} & 500 & 7 & 26 & 8.05 & 46.8 & 61.8 & 48.6 & 25.7 & 49.9 & 61.0 \\
        YOLOv12-S-turbo~\cite{tian2025yolov12} & 600 & 9 & 19 & 6.28  & 47.5 & 64.1 & - & - & - & - \\
        YOLO11-S~\cite{yolo11} & 500 & 9 & 22 & 7.05  & 46.6 & 63.4 & 50.3 & 28.7 & 51.3 & 64.1 \\
        D-FINE-S~\cite{peng2024d} & 120 & 10 & 25 & 3.66 & 48.5 & 65.6 & 52.6 & 29.1 & 52.2 & 65.4 \\
        {\ourmethod}-S~\cite{huang2025deim} & 120 & 10 & 25 & 3.66 & 49.0 & 65.9 & 53.1 & 30.4 & 52.6 & 65.7 \\
        YOLOv8-S~\cite{yolov8} & 500 & 11 & 29 & 6.97  & 44.9 & 61.8 & 48.6 & 25.7 & 49.9 & 61.0 \\
        \rowcolor[gray]{0.95} \textbf{{\ourmethod}v2-S} & 120 & 10 & 50 & 5.78 & 50.9 & 68.3 & 55.1 & 31.4 & 55.3 & 70.3 \\

        \midrule
        YOLOv10-M~\cite{wang2024yolov10} & 500 & 15 & 59 & 4.70 & 51.1 & 68.1 & 55.8 & 33.8 & 56.5 & 67.0 \\
        D-FINE-M~\cite{peng2024d} & 120 & 19 & 57 & 5.91  & 52.3 & 69.8 & 56.4 & 33.2 & 56.5 & 70.2 \\
        {\ourmethod}-M~\cite{huang2025deim} & 90 & 19 & 57 & 5.91 & 52.7 & 70.0 & 57.3 & 35.3 & 56.7 & 69.5 \\
        RT-DETRv2-S~\cite{lv2024rt} & 120 & 20 & 60 & 4.61  & 48.1 & 65.1 & 57.4 & 36.1 & 57.9 & 70.8 \\
        YOLOv12-M-turbo~\cite{tian2025yolov12} & 600 & 20 & 60 & 8.44 & 52.6 & 69.5 & - & - & - & - \\
        YOLO11-M~\cite{yolo11} & 500 & 20 & 68 & 9.02 & 51.2 & 67.9 & 55.3 & 33.0 & 56.7 & 67.5 \\
        YOLOv9-M~\cite{wang2024yolov9} & 500 & 20 & 76 & 10.18  & 51.4 & 67.2 & 54.6 & 32.0 & 55.7 & 66.4 \\
        Gold-YOLO-S~\cite{wang2024gold} & 300 & 22 & 46 &  6.96 & 46.4 & 63.4 & - & 25.3 & 51.3 & 63.6 \\
        \rowcolor[gray]{0.95} \textbf{{\ourmethod}v2-M} & 90 & 18 & 84 & 8.80 & 53.0 & 70.2 & 57.6 & 34.2 & 57.4 & 71.5 \\

        \midrule
        YOLOv10-L~\cite{wang2024yolov10} & 500 & 24 & 120 & 7.38 & 53.2 & 70.1 & 58.1 & 35.8 & 58.5 & 69.4 \\
        YOLO11-L~\cite{yolo11} & 500 & 25 & 87 & 10.41 & 53.4 & 70.1 & 58.2 & 35.6 & 59.1 & 69.2 \\
        YOLOv9-C~\cite{wang2024yolov9} & 500 & 25 & 102 & 10.76 & 53.0 & 70.2 & 57.8 & 36.2 & 58.5 & 69.3 \\
        YOLOv8-M~\cite{yolov8} & 500 & 26 & 79 & 9.46 & 50.2 & 67.2 & 54.6 & 32.0 & 55.7 & 66.4 \\
        YOLOv12-L-turbo~\cite{tian2025yolov12} & 600 & 27 & 82 & 10.45 & 54.0 & 70.6 & - & - & - & - \\
        YOLOv10-X~\cite{wang2024yolov10} & 500 & 30 & 160 & 10.47 & 54.4 & 71.3 & 59.3 & 37.0 & 59.8 & 70.9 \\
        D-FINE-L~\cite{peng2024d} & 72 & 31 & 91 & 8.15 & 54.0 & 71.6 & 58.4 & 36.5 & 58.0 & 71.9 \\
        \ourmethod-L~\cite{huang2025deim} & 50 & 31 & 91 & 8.15 & 54.7 & 72.4 & 59.4 & 36.9 & 59.6 & 71.8 \\
        RT-DETRv2-M~\cite{lv2024rt} & 120 & 31 & 92 & 6.91  & 49.9 & 67.5 & 58.6 & 35.8 & 58.6 & 72.1 \\
        Gold-YOLO-M~\cite{wang2024gold} & 300 & 41 & 88 & 9.21 & 51.1 & 68.5 & - & 32.3 & 56.1 & 68.6 \\
        RT-DETRv2-L~\cite{lv2024rt} & 72 & 42 & 136 & 9.29 & 53.4 & 71.6 & 57.4 & 36.1 & 57.9 & 70.8 \\
        YOLOv8-L~\cite{yolov8} & 500 & 43 & 165 & 12.20 & 52.9 & 69.8 & 57.5 & 35.3 & 58.3 & 69.8 \\
        \rowcolor[gray]{0.95} \textbf{\ourmethod v2-L} & 60 & 32 & 145 & 10.47 & 56.0 & 73.4 & 60.9 & 37.5 & 60.8 & 75.2 \\

        \midrule
        YOLOv9-E~\cite{wang2024yolov9} & 500 & 57 & 189 &  20.52 & 55.6 & 72.8 & 60.6 & 40.2 & 61.0 & 71.4 \\
        YOLO11-X~\cite{yolo11} & 500 & 57 & 195 & 15.53 & 54.7 & 71.6 & 59.5 & 37.7 & 59.7 & 70.2 \\
        YOLOv12-X-turbo~\cite{tian2025yolov12} & 600 & 59 & 185 & 15.79 & 55.7 & 72.2 & - & - & - & - \\
        D-FINE-X~\cite{peng2024d} & 72 & 62 & 202 & 12.90 & 55.8 & 73.7 & 60.2 & 37.3 & 60.5 & 73.4 \\
        \ourmethod-X~\cite{huang2025deim} & 50 & 62 & 202 & 12.90 & 56.5 & 74.0 & 61.5 & 38.8 & 61.4 & 74.2 \\
        YOLOv8-X~\cite{yolov8} & 500 & 68 & 257 & 15.89 & 53.9 & 71.0 & 58.7 & 35.7 & 59.3 & 70.7 \\
        Gold-YOLO-L~\cite{wang2024gold} & 300 & 75 & 152 & 12.31 & 53.3 & 70.9 & - & 33.8 & 58.9 & 69.9 \\
        RT-DETRv2-X~\cite{lv2024rt} & 72 & 76 & 259 & 13.88 & 54.3 & 72.8 & 58.8 & 35.8 & 58.8 & 72.1 \\
        \rowcolor[gray]{0.95} \textbf{\ourmethod v2-X} & 50 & 50 & 200 & 13.75 & 57.8 & 75.4 & 63.2 & 39.2 & 62.9 & 75.9  \\
        \bottomrule
    \end{tabular}
    \end{adjustbox}
    \end{center}
\end{table*}

\begin{table*}[t]
    \centering
    \caption{\textbf{Comparison with real-time object detectors on COCO~\cite{lin2014microsoft} \texttt{val2017} for ultra-light models}.}
    
    \label{tab:my_sorted_label_asc}
    \begin{tabular}{l c c c c}
        \toprule
        \hline
        Model & Input Size & Params (M) & FLOPs (G) & COCO AP \\
        \midrule
        \hline
        NanoDet-M\cite{lyu2021nanodetplus} & 416x416 & 1.0 & 0.7 & 23.5 \\
        \rowcolor[gray]{0.95} \textbf{DEIMv2-Atto} & 320x320 & 0.5 & 0.8 & \textbf{23.8} \\
        \midrule
        YOLOX-Nano\cite{zheng2021yolox} & 416x416 & 0.9 & 1.1 & 25.8 \\
        PicoDet-S\cite{yu2021pp} & 416x416 & 1.0 & 1.2 & 30.7 \\
        PP-YOLO-Tiny\cite{long2020pp} & 416x416 & 1.1 & 1.0 & 22.7 \\
        PicoDet-ShufflenetV2 1x\cite{yu2021pp} & 416x416 & 1.2 & 1.5 & 30.0 \\
        \rowcolor[gray]{0.95} \textbf{DEIMv2-Femto} & 416x416 & 1.0 & 1.7 & \textbf{31.0} \\
        \midrule
        YOLOv10-N\cite{wang2024yolov10} & 640x640 & 2.3 & 6.7 & 38.5 \\
        YOLOv8-N\cite{yolov8} & 640x640 & 3.2 & 8.7 & 37.4 \\
        YOLOv6-3.0-N\cite{li2023yolov6} & 640x640 & 4.7 & 11.4 & 37.0 \\
        \rowcolor[gray]{0.95} \textbf{DEIMv2-Pico} & 640x640 & 1.5 & 5.2 & \textbf{38.5} \\
        \midrule
        YOLOv12-N~\cite{tian2025yolov12} & 640x640 & 2.6 & 6.5 & 40.6 \\
        {\ourmethod}-Nano~\cite{huang2025deim} & 640x640 & 3.6 & 6.9 & 43.0 \\
        D-FINE-Nano~\cite{peng2024d} & 640x640 & 3.8 & 7.2 & 42.8 \\
        \rowcolor[gray]{0.95} \textbf{{\ourmethod}v2-Nano} & 640x640 & 3.6 & 6.9 & \textbf{43.0} \\
        \bottomrule
    \end{tabular}
\end{table*}

\paragraph{Efficient Decoder.} 
We enhance the standard deformable attention decoder~\cite{zhu2020deformable} by incorporating several efficiency-oriented techniques widely adopted in the Transformer community, achieving a favorable performance–cost trade-off. Specifically, we integrate SwiGLUFFN~\cite{shazeer2020glu} to strengthen nonlinear representation capacity, RMSNorm~\cite{zhang2019root} to stabilize and accelerate training efficiently. Noting that object query locations undergo minimal changes during iterative refinement, we further propose sharing a single position embedding across all decoder layers, eliminating redundant computation.

\paragraph{Enhanced Dense O2O.}
In our previous DEIM~\cite{huang2025deim}, we proposed Dense O2O, which increases the number of objects per training image to provide stronger supervision, improving convergence and performance. Its effectiveness was initially demonstrated using image-level augmentations such as Mosaic and MixUp~\cite{zhang2017mixup}. In DEIMv2, we further explore Dense O2O at the object level with Copy-Blend, which adds new objects without their backgrounds. Unlike Copy-Paste~\cite{ghiasi2021simple}, which fully overwrites the target region, Copy-Blend blends new objects with the image, better suiting our scenario and consistently improving performance.

\paragraph{Training setting and loss.} The overall optimization objective is a weighted sum of five components: Matchability-Aware Loss (MAL)~\cite{huang2025deim}, Fine-Grained Localization (FGL) Loss~\cite{peng2024d}, Decoupled Distillation Focal (DDF) Loss~\cite{peng2024d}, BBox Loss (L1), and GIoU Loss~\cite{rezatofighi2019generalized}. The total loss is defined as:
\begin{equation}
    \begin{aligned}
    L_{total} = & \lambda_1 L_{mal} + \lambda_2 L_{fgl} + \lambda_3 L_{ddf} \\
    & + \lambda_4 L_{bbox} + \lambda_5 L_{giou}
    \end{aligned}
\end{equation}
with weights set to $\lambda_1 = 1.0$, $\lambda_2 = 0.15$, $\lambda_3 = 1.5$, $\lambda_4 = 5$, and $\lambda_5 = 2$ across all experiments. 

We summarize the training hyperparameters in Table~\ref{tab:hyperparams}, covering input resolution, learning rate, training epochs, and Dense O2O settings. An interesting observation is that applying FGL and DDF losses to ultra-lightweight models degrades performance. We attribute this to their limited capacity and inherently weaker baseline accuracy, which reduces the effectiveness of self-distillation. Consequently, we exclude these two components (i.e., the local loss) from training the Pico, Femto, and Atto variants.

\section{Experiments}

\paragraph{Comparison to state-of-the-art real-time object detectors.} Table~\ref{tab:XLMS} summarizes the performance of DEIMv2 across the S, M, L, and X variants, demonstrating substantial improvements over prior state-of-the-art detectors. For instance, the largest variant, DEIMv2-X, attains 57.8 AP with $\sim$50M parameters and 200 GFLOPs, surpassing the previous best DEIM-X (56.5 AP with 62M parameters and 202 GFLOPs). At the smaller end, DEIMv2-S sets a new milestone as the first model with fewer than 10M parameters to exceed the 50 AP threshold on COCO. This marks a clear improvement over the prior DEIM-S (49.0 AP with 10M parameters), while requiring nearly the same model size. While CNN-based backbones are generally more hardware-friendly, our ViT-based backbone achieves a lightweight design with fewer parameters, offering better scalability and deployment flexibility. It is worth noting that the latency of the proposed methods has not been optimized. Techniques such as Flash Attention~\cite{dao2022flashattention}, as in Yolov12~\cite{tian2025yolov12}, could further accelerate inference.

Interestingly, when comparing DINOv3-based DEIMv2 models with their previous DEIM counterparts under comparable parameter and FLOP budgets, the accuracy gains primarily arise from improvements on medium and large objects, while performance on small objects remains largely unchanged. For instance, DEIMv2-S achieves 55.3 AP$_M$ and 70.3 AP$_L$, clearly surpassing DEIM-S (52.6 AP$_M$ and 65.7 AP$_L$), yet the small-object scores are nearly identical (31.4 vs. 30.4 AP$_S$). A similar trend is observed for larger models: DEIMv2-X improves AP$_M$ from 61.4 to 62.8 and AP$_L$ from 74.2 to 75.9, while its small-object AP (39.2) remains close to that of DEIM-M (38.8). These results indicate that DEIMv2’s main advantage lies in enhancing the representation and detection of medium-to-large objects, whereas small-object detection remains a challenge across scales. This observation further confirms that DINOv3 excels at capturing strong global semantics but has limited ability to represent fine-grained details. Exploring ways to better integrate DINOv3 features into real-time detectors thus represents an interesting direction for future work.

\paragraph{Comparison to competitive ultra-light object detectors.} The ultra-light variants of DEIMv2 also exhibit strong performance, as summarized in Table~\ref{tab:my_sorted_label_asc}. DEIMv2-Atto, with only 0.49M parameters, achieves performance comparable to NanoDet-M despite its substantially smaller size. Similarly, DEIMv2-Pico attains performance on par with YOLOv10-N~\cite{wang2024yolov10} while requiring less than half the parameters. These results underscore the effectiveness of DEIMv2 in extremely compact regimes and highlight its suitability for deployment on resource-constrained edge devices.

\section{Conclusion}

In this report, we introduced DEIMv2, a new generation of real-time object detectors that combines the strong semantic representations of DINOv3 with our lightweight STA. Through careful design and scaling, DEIMv2 achieves state-of-the-art performance across the full spectrum of model sizes. At the high end, DEIMv2-X delivers 57.8 AP with significantly fewer parameters than previous large-scale detectors. At the compact end, DEIMv2-S is the first model of its size to surpass 50 AP, and the ultra-lightweight DEIMv2-Pico matches YOLOv10-N while using over 50\% fewer parameters. Together, these results demonstrate that DEIMv2 is not only efficient but also highly scalable, offering a unified framework that advances the accuracy–efficiency frontier. This versatility makes DEIMv2 well-suited for deployment in diverse scenarios, ranging from resource-constrained edge devices to high-performance detection systems, paving the way for broader adoption of real-time detection in practical applications.

{
    \small
    \bibliographystyle{ieeenat_fullname}
    \bibliography{main}
}

\end{document}